\documentclass{article}

\usepackage{natbib}
\usepackage[final]{neurips_ts4h_2022}

\usepackage{graphicx}
\usepackage{subcaption}

\usepackage[utf8]{inputenc} 
\usepackage[T1]{fontenc}    
\usepackage{hyperref}       
\usepackage{url}            
\usepackage{booktabs}       
\usepackage{amsfonts}       
\usepackage{nicefrac}       
\usepackage{microtype}      
\usepackage{xcolor}         

\title{sEHR-CE: Language modelling of structured EHR data for efficient and generalizable patient cohort expansion}

\author{%
  Anna Munoz-Farre\\
  BenevolentAI\\
  \texttt{anna.munoz.farre@benevolent.ai} \\
   \And
   Harry Rose\\
     BenevolentAI\\
   \texttt{harry.rose@benevolent.ai} \\
  \AND
   Sera Aylin Cakiroglu\\
  BenevolentAI\\
  \texttt{aylin.cakiroglu@benevolent.ai} \\
}

\begin{document}

\maketitle

\begin{abstract}
Electronic health records (EHR) offer unprecedented opportunities for in-depth clinical phenotyping and prediction of clinical outcomes. Combining multiple data sources is crucial to generate a complete picture of disease prevalence, incidence and trajectories. The standard approach to combining clinical data involves collating clinical terms across different terminology systems using curated maps, which are often inaccurate and/or incomplete. Here, we propose sEHR-CE, a novel framework based on transformers to enable integrated phenotyping and analyses of heterogeneous clinical datasets without relying on these mappings. We unify clinical terminologies using textual descriptors of concepts, and represent individuals’ EHR as sections of text. We then fine-tune pre-trained language models to predict disease phenotypes more accurately than non-text and single terminology approaches. We validate our approach using primary and secondary care data from the UK Biobank, a large-scale research study. Finally, we illustrate in a type 2 diabetes use case how sEHR-CE identifies individuals without diagnosis that share clinical characteristics with patients.
\end{abstract}

\section{Introduction}
\label{sec:intro}

Electronic health records (EHR) are collected as part of routine medical care, and include demographic information, disease diagnoses, laboratory results, medication prescriptions, etc., providing  a patient’s clinical state over time. Recent machine learning techniques have been used to exploit the richness of EHR data at scale for diagnosis, prognosis, treatment and understanding of disease \citep{https://doi.org/10.48550/arxiv.1907.09538, STEINBERG2021103637}. Many medical terminologies are used across clinical data sets, and the standard practice involves mapping clinical terms across different resources \citep{https://doi.org/10.48550/arxiv.1907.09538, HASSAINE2020103606} or onto common data models \citep{10.7326/0003-4819-153-9-201011020-00010}. 

Here, we propose sEHR-CE (language modelling of \textbf{S}tructured \textbf{EHR} data for patient \textbf{C}ohort \textbf{E}xpansion), a novel framework based on transformers to enable the integrated analysis of multiple clinical resources without relying on any manual curation and mapping. Using text descriptions of concepts as input, our method generalises across data modalities and terminologies, i.e. text and structured EHR. This enables us to leverage a plethora of pre-trained language models like PubMedBERT \citep{pubmedbert} to encode each patient’s medical record. We ask the model to learn representations of clinical histories from diagnosed patients (cases) to predict phenotypes (such as diseases). In the absence of a directly comparable model, we evaluate the performance of our model against that of  \citet{https://doi.org/10.48550/arxiv.1907.09538}, as the state-of-the-art approach for learning clinical term embeddings for future disease prediction. See Appendix \ref{sec:related_work} for a more complete survey. 

In our validation experiments, we will show that the model can then identify individuals with missing diagnoses (controls) that share a similar clinical history with cases, indicating they might have the disease or be at risk of developing it. The contributions of our paper are \textbf{(i)} the presentation of a cohort expansion method that provides phenotype predictions outperforming non-text and single terminology approaches, and \textbf{(ii)} an in-depth qualitative evaluation demonstrating that positively predicted controls share similar clinical representations with cases, providing a high degree of evidence that these may be previously undiagnosed or misdiagnosed individuals. Finally, we demonstrate our method’s potential for patient stratification by disease severity.

\section{Methods}
\label{sec:method}

\textbf{Input Generation}. For each EHR source and associated ontology $a \in \mathcal{A}$, we denote the set of concepts (e.g. clinical terms) as $\Theta_a$, and the set of text descriptions as $\Xi_a$. The total vocabulary of concepts and text descriptions across all ontologies is denoted by $\Theta = \bigcup\limits_{a \in \mathcal{A}} \Theta_a$ and $\Xi = \bigcup\limits_{a \in \mathcal{A}} \Xi_a$, respectively. For each patient, we define their full clinical history through time and across sources by the concatenation of sequences of clinical descriptions $(\xi_{\theta_1}, \ldots ,\xi_{\theta_t})$, $\xi_{\theta_i} \in \Xi$, $i=1, \ldots, t$, ordered in time. More details and an example can be found in Appendix \ref{fusing}, and Figure \ref{figure:hosp_gp}. To form the input, we process the raw text sequence of descriptions into tokens (e.g. words and sub-words) $X = W(\xi_{\theta_1},\ldots,\xi_{\theta_t}) = (x_1,\ldots,x_{n})$ under a fixed size vocabulary $V$ with the WordPiece tokenizer $W$ \citep{https://doi.org/10.48550/arxiv.1609.08144}.

\begin{figure}[ht]
\centering
\includegraphics[width=10cm]{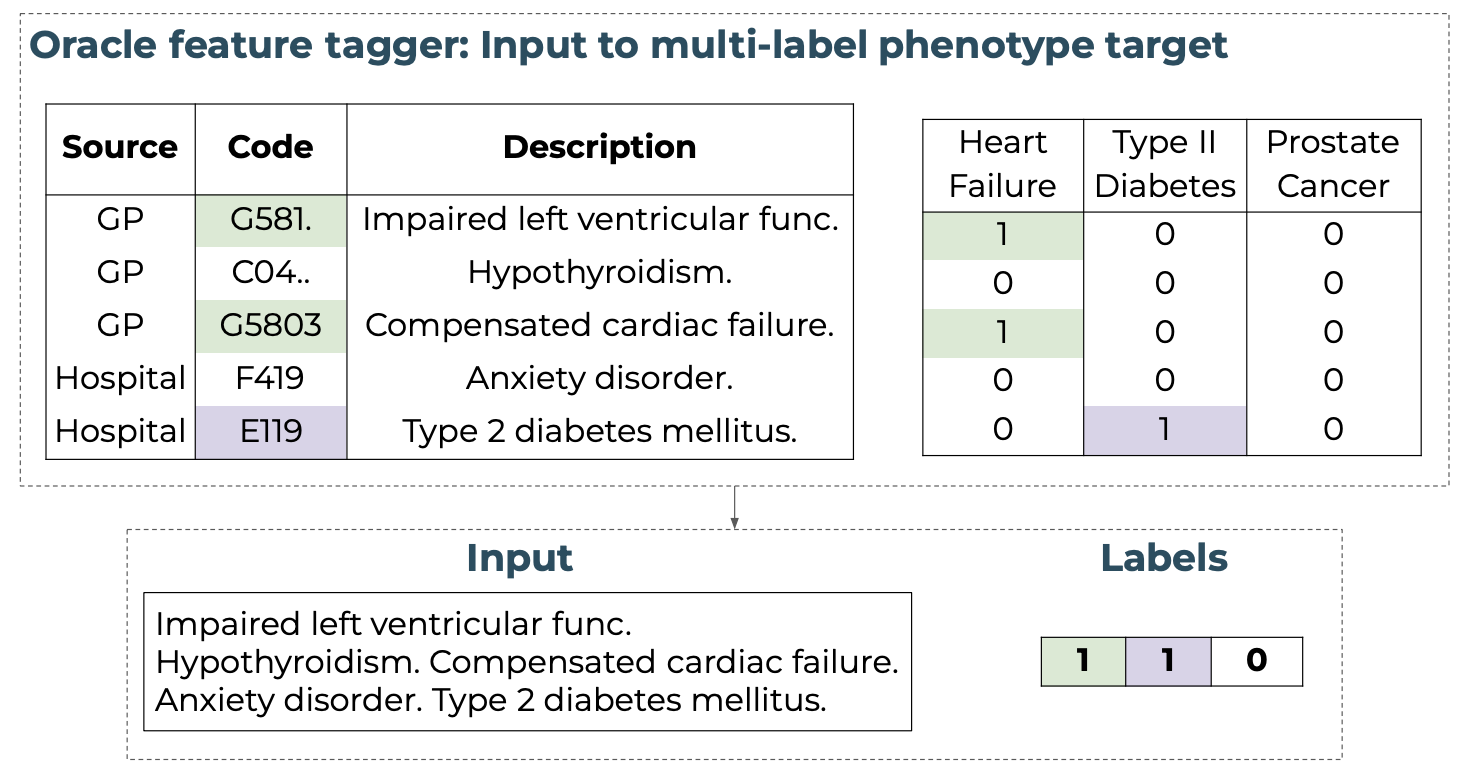}
\caption{Schematic of mapping a unified clinical history to an input text sequence with multi-label target vector using an oracle annotation \protect\footnotemark.} 
\label{figure:labelling} 
\end{figure}

\textbf{Label Generation}. Let $\Delta=(d_1, \ldots , d_D)$ denote an ordered set of clinical outcomes or events, in our case disease phenotypes $d_i$. For each phenotype $d_i\in\Delta$, we let $\mathbf{1}_{d_i}$ be an indicator function that assigns a binary label to individuals according to the presence or absence of $d_i$. To define $\mathbf{1}_{d_i}$, we rely on external oracles or phenotype definitions, such as CALIBER (Appendix \ref{labelling}). Figure \ref{figure:labelling} shows a schematic of input and label generation for a given patient. 

\footnotetext{Any patient data shown is simulated and does not represent data of real patients.}

\textbf{Model Design}. Let $X^{(p)} = (x^{(p)}_1, \ldots ,x^{(p)}_{n})$ denote the tokenized input sequence of individual $p$. It forms the input to an encoding function
$\mathbf{x}^{(p)}_1, \ldots, \mathbf{x}^{(p)}_n$ = \textit{Encoder}$(X^{(p)})$, where each $\mathbf{x}_i$ is a fixed-length vector representation of each input token $x_i$. Let $\mathbf{y}^{(p)}=(y^{(p)}_1, \ldots, y^{(p)}_D)$, $y^{(p)}_i\in \{0,1\}$, be labels denoting presence or absence of phenotypes $d_1,\ldots, d_D$. Given a learned representation over inputs, we decode over $\mathbf{y}^{(p)}$ under the predictive model $\mathbf{P}(\mathbf{y}^{(p)}|X^{(p)})$. We calculate the probability of each phenotype $d_i$ given the input sequence encoding $\mathbf{P}(y^{(p)}_i|\textbf{x}^{(p)}_1,\dots, \textbf{x}^{(p)}_n)$ via a decoder module. Specifically, we decode the representation into logits per phenotype $z^{(p)}_1, \dots, z^{(p)}_D =$ \textit{Decoder}$(\textbf{x}^{(p)}_1,\dots, \textbf{x}^{(p)}_n)$ and calculate the probability per phenotype as $\mathbf{P}(y^{(p)}_d|\textbf{x}^{(p)}_1,\dots, \textbf{x}^{(p)}_n) = \sigma(z^{(p)}_d)$, where $\sigma$ denotes the sigmoid function (Figure \ref{figure:model_diagram_small}). We will omit the superscript $(p)$ denoting the sample index in the remainder of the text.

\subsection{Data Augmentation}\label{data_augmentation}
\begin{figure*}[h]
\setlength{\belowcaptionskip}{-5pt}
\centering
\includegraphics[width=14cm]{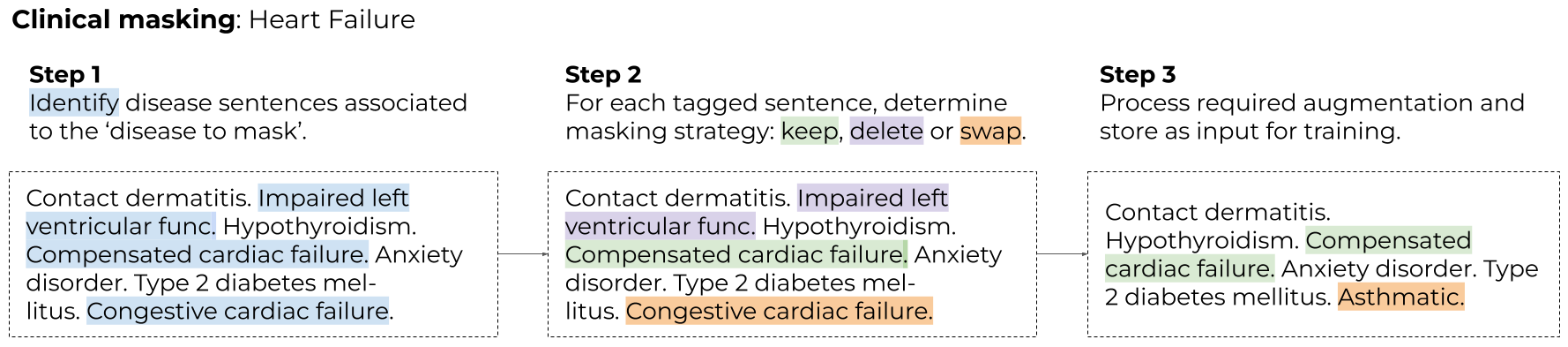}
\caption{Example of clinical masking strategy for a given patient history with \textbf{Heart Failure}.}
\label{figure:augment} 
\end{figure*}

\textbf{Clinical Masking}. We mask input descriptions $\xi_{\theta}$ from term-description pairs $(\theta, \xi_{\theta})$ with $\mathbf{1}_{d}(\theta, \xi_{\theta})=1$ for $d \in \Delta$ using the following clinical masking strategy during training and validation as in \citet{DBLP:journals/corr/abs-1810-04805, https://doi.org/10.48550/arxiv.1901.11196}: \textbf{Remove} $\xi_{\theta}$ with 80\% probability, \textbf{retain} $\xi_{\theta}$ with 
10\% probability, and \textbf{replace} $\xi_{\theta}$ with a randomly selected description from the corpus with 10\% probability.
Figure \ref{figure:augment} displays a worked example. During testing, these term-description pairs are fully removed from the input sequence.

\textbf{Comorbidities}. This masking approach is straightforward if an individual has only one positive label, but many people have comorbidities, e.g. co-occurring conditions. To allow for a sample with multiple positive labels $d_{i_1}, \ldots , d_{i_n}$, we create $n$ input samples by replicating both the input sequence and target vector of phenotype labels. Consider the $j$th replicated input sequence, and let $d = d_{i_j}$. We mask this replicated input sequence with the masking strategy for phenotype $d$ described in the previous section; e.g. $\xi_{\theta}$ is masked if  $\mathbf{1}_{d}(\theta, \xi_{\theta})=1$ (Figure \ref{figure:masking_comorbidities}).

\subsection{Defining a Loss Function in the Presence of Comorbidities and varying Prevalences}\label{loss_weights}
\begin{figure*}[h!]
\centering
\setlength{\belowcaptionskip}{-10pt}
\includegraphics[width=11cm]{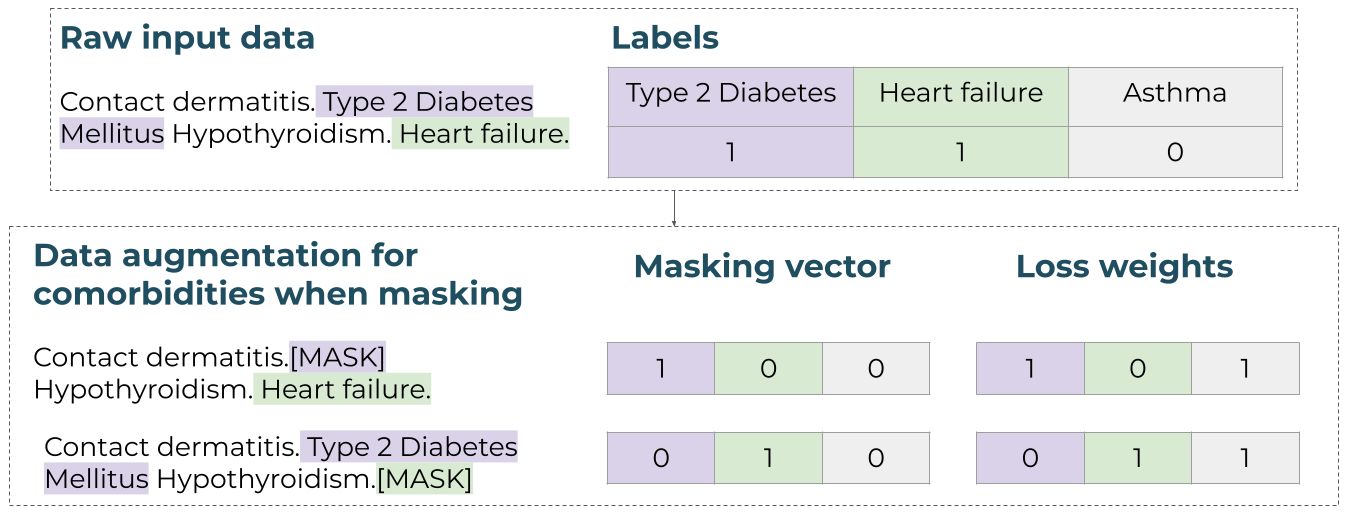}
\caption{ Example of an input sequence containing  descriptions relevant to two positive phenotype labels in the target vector $y$. The input sequence is duplicated, and each phenotype is masked once and a loss weights vector is defined. Clinical terms relating to a directly associated phenotype $d$ remain in the input data ($y_d=1$ and $\gamma_d=0$) but get assigned a loss weight $w_d=0$  and so do not contribute to the loss function.}
\label{figure:masking_comorbidities} 
\end{figure*}

\textbf{Loss weights}. The data augmentation approach for comorbidities may lead to our model overfitting when an input sequence contains the descriptions associated with an existing phenotype $d$ that is not masked. To account for the contribution of the target vector $\mathbf{y}$ to the loss function in these scenario, we define a binary masking vector $\mathbf{\gamma}^{j}=(\gamma^{j}_1, \ldots, \gamma^{j}_D)$ where $\gamma^{j}_j=1$ and $\gamma^{j}_k=0$, $k=1,\ldots,D$, $k\not=j$. Individuals with no positive phenotype labels are assigned an all-zero masking vector (Figure \ref{figure:masking_comorbidities}). Then, for a given input text sequence $X$, target label vector $\mathbf{y}=(y_1, \ldots, y_D)$ and masking vector $\mathbf{\gamma}=(\gamma_{1}, \ldots, \gamma_{D})$, we define the loss weights by setting $\omega_{d} = 0$ if $y_{d} = 1$ and $\gamma_d = 0$, and $\omega_{d} = 1$ if  $y_{d} = 0$ or $y_{d} = 1$ and $\gamma_d = 1$. 

\textbf{Positive weights}. Because some diseases are very rare, whereas others are very common, our prediction classes are expected to be highly imbalanced in the practical setting. For cohort expansion, we want to increase recall while balancing a decline in precision. For $d\in\Delta$, let $TN_d$ and $TP_d$ denote the total number of negative and positive examples of $d$ in the training set. We define the positive weight $\rho_d$ as $\rho_d=TN_d/TP_d$, for all d $\in\Delta.$ 

Our loss function is defined as a mean-reduced binary cross-entropy loss function over phenotypes, where disease prevalence and comorbidities are handled with loss and positive weights: 

\begin{equation}\label{lossfunction}
\mathcal{L}^{(p)}= -\frac{1}{|\Delta|} \sum_{d \in \Delta} \omega_{d}^{(p)} (\rho_dy_{d}^{(p)} \log \sigma({z_{d}^{(p)}}) \\ + (1-y_{d}^{(p)}) \log \sigma(1-z_{d}^{(p)})),
\end{equation}
where $\sigma$ denotes the sigmoid function, $\omega_d$ the comorbidity-derived loss weight, $\rho_d$ the positive weight, and $z_{d}^{(p)}$ the predicted probability for phenotype $d\in \Delta$ for sample (e.g. individual) $p$. 

\section{Experiments and Results}
\label{sec:results}
 
\textbf{Data.} This research has been conducted using the UK Biobank (UKBB) Resource under Application Number 43138, a large-scale research study of around $500$k individuals \citep{10.1371/journal.pmed.1001779}. We restrict the data set to those that have entries in both hospital and GP records, reducing our cohort to $155$k. To assess the quality of our model predictions, we choose four diseases that differ in terms of prevalence and clinical characteristics (Appendix \ref{apd:model}): Type 2 diabetes mellitus (T2DM), Heart failure (HF), Breast Cancer, and Prostate Cancer. We use validated phenotype definitions from CALIBER \citet{kuan2019chronological} to label patients with each of the diseases (Appendix \ref{apd:data_proc}). We test the performance of our model on its ability to diagnose known cases, compare it to other methods, and evaluate associations of clinical features with predictions on T2DM. 

\textbf{sEHR-CE.} We use the pre-trained language model PubMedBERT \citep{pubmedbert} as the encoder of the tokenised input sequences of clinical term descriptions. Since our input systematically differs from the general scientific text on which PubMedBERT was trained, we fine-tune on the masked-language modeling (MLM) task using the full UKBB cohort. The proposed model sEHR-CE uses the fine-tuned encoder and a fully connected linear layer as the decoder. To train on the multi-label classification task of outcome prediction, we split our data set into five equally sampled folds \textit{$f_0$,...$f_4$} containing unique patients, and mask the data according to our strategy (Section \ref{data_augmentation}, Figure \ref{figure:input_diagram}). We train a total of five models on three folds, holding back folds $f_i$ for validation and $f_{(i+1)\bmod{5}}$ for testing for model $i$, $i=1,\ldots,5$ (Figure \ref{figure:folds_diagram}). All results presented are predictions on the independent test set. For masking and training details, see Appendix \ref{apd:training}.

\textbf{Baseline Models}. We compare the performance of our model sEHR-CE to BEHRT \citep{https://doi.org/10.48550/arxiv.1907.09538}, which takes a tokenised sequence of clinical terms, age and position embeddings as input. Ontologies from hospital and GP records are mapped to CALIBER definitions, removing unmapped terms (more details in Appendix \ref{apd:eval}). We train five such BEHRT models to predict an individual developing the four phenotypes on the same five data splits as sEHR-CE. Similarly, we train five sEHR-CE models restricted to CALIBER tokens (denoted sEHR-CE-codes) for comparison. Figure \ref{figure:disease_dist_all} shows the distributions of predicted probabilites for all phenotypes across all methods. sEHR-CE shows the best performance across all four phenotypes in terms of recall at $0.5$ and AUC on the test set (Table \ref{table:results_SOTA}, Figure \ref{figure:auprc}). BEHRT performs slightly better than sEHR-CE-codes, indicating a benefit of adding visit position and age. Performance varies across phenotypes, presumably due to different clinical characteristics making some diseases easier to predict than others.

\begin{table*}[h]
\centering
\small
\addtolength{\tabcolsep}{-1pt}
\begin{tabular*}{\textwidth}{l|clclclclcl}
\hline
              & \multicolumn{2}{c}{T2D}            & \multicolumn{2}{c}{HF}             & \multicolumn{2}{c}{Breast Cancer}  & \multicolumn{2}{c}{Prostate Cancer} & \multicolumn{2}{c}{Average}        \\
              & \multicolumn{1}{l}{Recall} & PRC & \multicolumn{1}{l}{Recall} & PRC & \multicolumn{1}{l}{Recall} & PRC & \multicolumn{1}{l}{Recall}  & PRC & \multicolumn{1}{l}{Recall, std} & PRC \\ \hline
sEHR-CE-codes    & 0.64                       & 0.70  & 0.76                       & 0.60  & 0.42                       & 0.45  & 0.36                        & 0.26  & 0.55 $\pm$ 0.19                       & 0.50  \\
BEHRT          & 0.66                       & 0.70  & 0.78                       & 0.61  & 0.48                       & 0.53  & 0.41                        & 0.29  & 0.58 $\pm$ 0.17                      & 0.53  \\
\textbf{sEHR-CE} & \textbf{0.74}              & 0.74  & \textbf{0.85}              & 0.69  & \textbf{0.55}              & 0.55  & \textbf{0.57}               & 0.47  & \textbf{0.68 $\pm$ 0.14}              & 0.61  \\ \hline
\end{tabular*}
\caption{Average and phenotype specific recall and AUCPR at $0.5$ on the test sets.}
\label{table:results_SOTA} 
\end{table*}

\subsection{Clinical evaluation on Type 2 Diabetes Mellitus}

T2DM lends itself as a use case to qualitatively evaluate the predictions of missed cases, as it is a well-studied, slowly developing disease with varying disease severity. We used thresholds based on the 98$^{th}$, 90$^{th}$ and 12$^{th}$ percentiles of sEHR-CE's predicted probabilities (Figure \ref{figure:disease_dist_all}) to define five different groups (Table \ref{table:patient_groups}): regular cases, cases predicted with high probability, cases predicted with low probability, regular controls and controls predicted with high probability (missed cases). 
 


\begin{figure}
    \centering
    \begin{subfigure}[t]{0.49\textwidth}
        \includegraphics[width=\textwidth]{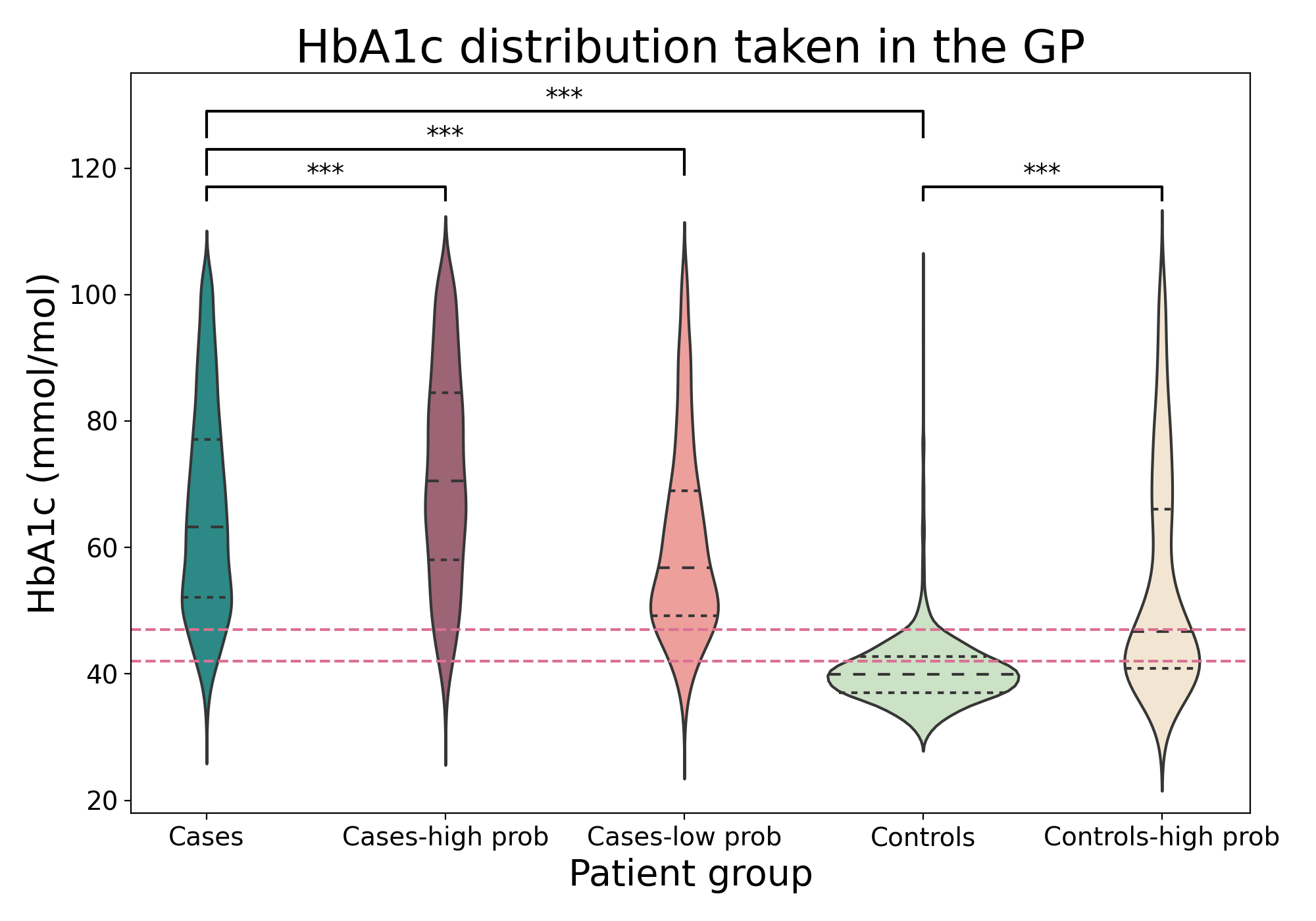}
    \caption{}
    \label{figure:hba1c}
    \end{subfigure}
    \begin{subfigure}[t]{0.49\textwidth}
        \includegraphics[width=\textwidth]{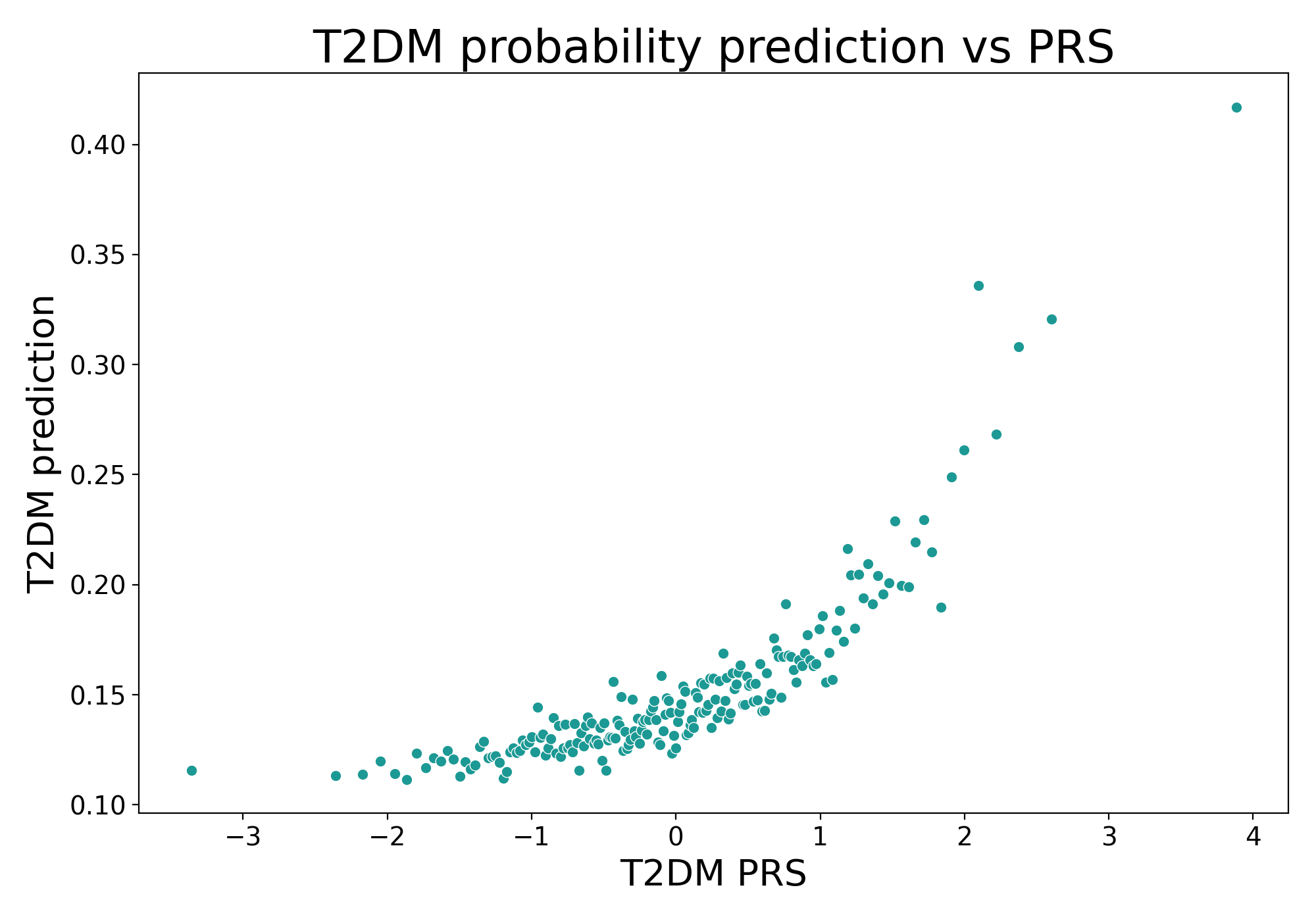}
        \caption{}
        \label{figure:prs}
    \end{subfigure}
    \caption{\ref{figure:hba1c} HbA1c distribution across groups in the test sets with diabetes diagnosis ranges shown as dashed lines. Significant p-values are indicated with *** (t-test, $\alpha = .001$).  \ref{figure:prs} Median T2DM prediction of individuals in the test sets grouped by the percentiles of the polygenic risk score.}
\end{figure}

\textbf{Measures of Disease Severity}. Haemoglobin A1c (HbA1c) is a blood biomarker used to diagnose and monitor diabetes. A level of 48mmol/mol or higher is considered diabetes; while a value between 42 and 48 mmol/mol is considered pre-diabetes. The input data did not include such biomarker data, so we use it here for evaluation. 
We aggregated HbA1c measurements taken in primary care (GP) of each individual by taking the $95$-th percentile value. Figure \ref{figure:hba1c} shows that cases predicted with high probability had the highest HbA1c mean levels. Cases identified with low probability were in the prediabetic range of HbA1c levels, indicating a less severe state. Missed cases had elevated HbA1c levels close to the prediabetic stage, representing individuals at risk of developing T2DM. We further investigated the association of the T2DM predicted probabilities with other measures of disease severity, expanded in Appendix \ref{apd:t2dm_eval}. Taken together, our results demonstrate that sEHR-CE's predicted probabilities of being diagnosed with T2DM are associated with disease severity.



\textbf{Polygenic risk scores}. Genetic risk for complex diseases like T2DM arises from many genetic changes that, when taken together, can increase an individual's risk of developing the disease, which can be defined by polygenic risk scores (PRS).  \cite{Sinnott-Armstrong660506} developed PRS  based on the UK Biobank participants for a set of diseases, including T2DM. We computed and standardised T2DM PRS across all individuals in our cohort \citep{prs_use}.  Figure \ref{figure:prs}  demonstrates that higher predicted probability of T2DM was associated with a higher genetic risk. 




\section{Conclusion}\label{sec:conclusion}
We presented a data-driven method for cohort expansion based on language modelling. Our approach fuses primary and secondary care data via text, and we propose a data augmentation approach to allow for comorbidities in a patient's history. Our method predicts disease phenotype labels more accurately than non-text and single terminology approaches. We presented a high degree of evidence that our model identifies previously undiagnosed individuals that can extend the original cohort for downstream analysis. Future work will consider methods that are less restrictive on sequence length \citep{transformer, longformer} and allow for irregular time steps \citep{shukla2021multitime} and age \citep{time2vec}, as well as adding more data sources (e.g. medications). 

\section{Acknowledgements}

This research has been conducted using the UK Biobank Resource under Application Number 43138. Using real patient data is crucial for clinical research and to find the right treatment for the right patient. We would like to thank all participants who are part of the UK Biobank, who volunteered to give their primary and secondary care and genotyping data for the purpose of research. UK Biobank is generously supported by its founding funders the Wellcome Trust and UK Medical Research Council, as well as the British Heart Foundation, Cancer Research UK, Department of Health, Northwest Regional Development Agency and Scottish Government.

We are particularly grateful to Prof. Spiros Denaxas, and Drs. Aaron Sim, Andrea Rodriguez-Martinez, Nicola Richmond, Sam Abujudeh, and Julien Fauqueur for their feedback, insightful comments and the many inspiring conversations. 

\bibliography{neurips}

\newpage
\appendix

\counterwithin{figure}{section}
\counterwithin{table}{section}

\section{Related Work}
\label{sec:related_work}
ML approaches have been applied to EHR data either using a cross-sectional matrix of diagnosis terms (e.g. an ICD-10 term)  \citep{8031150,Miotto2016} or sequential data as input \citep{DBLP:journals/corr/ChoiBS15, https://doi.org/10.48550/arxiv.1602.00357,NEURIPS2019_1d0932d7, 10.1145/2939672.2939823, DBLP:journals/corr/abs-1811-11005, DBLP:journals/corr/abs-1806-02873, DBLP:journals/corr/abs-1908-03971}.  
In the former, methods are blind to the order in which diagnoses occur and subsequently how a patient's disease profile develops. In the latter, most methods do not learn embeddings for the full sequence of diagnoses in a patient's medical history, and instead learn embeddings per diagnosis term or, at most, a short sequence of terms whether they are utilising LSTMs \citep{DBLP:journals/corr/ChoiBS15}, RNNs \citep{https://doi.org/10.48550/arxiv.1602.00357}, CNNs \citep{https://doi.org/10.48550/arxiv.1607.07519}, or transformer models \citep{https://doi.org/10.48550/arxiv.1907.09538}. To be able to deal with heterogeneous ontologies from different EHR data sources, all of these models rely on noisy and often lossy mappings across ontologies or on phenotyping algorithms, eg. manually curated groupings of ontology terms, such as CALIBER \citep{kuan2019chronological}. 

In contrast, our method uses the textual description of terms to learn representations across the full sequence of diagnoses in a patient's history. Recent work by \citet{pmlr-v174-hur22a} aims to unify EHR records by learning description-based embeddings from multiple data sources. Our work was developed in parallel independently and addresses the specific use case of cohort expansion, instead of merely providing examples of potential downstream applications to assess improvements of predictive power.

\section{Method details and figures}

\subsection{Fusing ontologies via text}\label{fusing}

We consider all EHR data sources with their ontologies where each concept has a textual descriptor. For example, the ontologies of GP and hospital records are made up of clinical terms (e.g. Read version 2/ Clinical Terms Version 3 and ICD9/ICD10 codes, respectively) and their description. For each EHR  source and associated ontology $a \in \mathcal{A}$, we denote the set of concepts (e.g. clinical terms in the case of GP or hospital records) within this ontology as $\Theta_a$ and the set of text descriptions as $\Xi_a$. The total vocabulary of concepts and text descriptions across all ontologies is denoted by $\Theta = \bigcup\limits_{a \in \mathcal{A}} \Theta_a$ and $\Xi = \bigcup\limits_{a \in \mathcal{A}} \Xi_a$, respectively. For each patient, we define their full clinical history through time and across sources as the sequence of time-indexed concepts as $(\theta_1,\dots, \theta_{t})$, $\theta_i \in \Theta$, $i=1, \ldots, t$.

\begin{figure}[h]
\centering
\setlength{\belowcaptionskip}{-10pt}
\includegraphics[width=13cm]{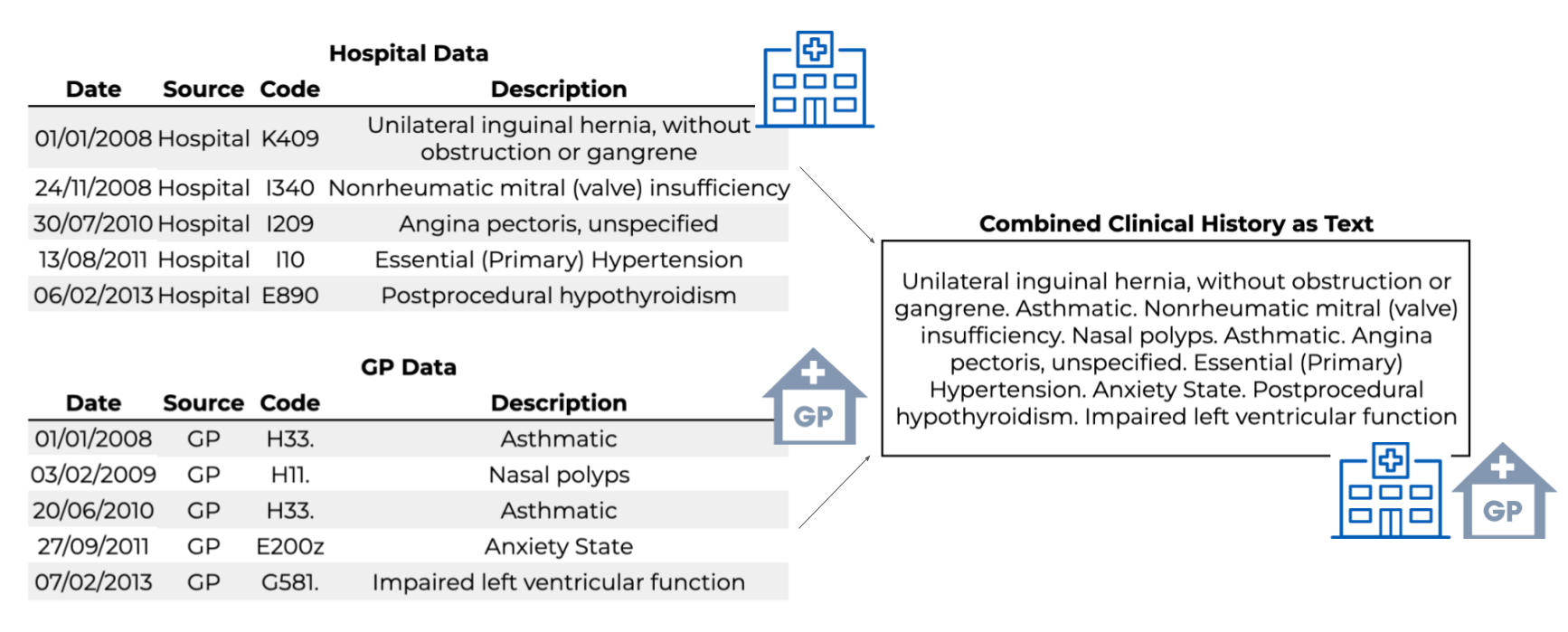}
\caption{Fusing primary (GP) and secondary (hospital) data into a single paragraph\protect\footnotemark.}
\label{figure:hosp_gp} 
\end{figure}

The premise of our work relies on the assumption that for every concept $\theta \in \Theta,$ there exists a unique text description $\xi_{\theta} \in \Xi$. For example, under the ICD-10 terminology, the alphanumeric code E11.9 has the associated description ‘Type 2 diabetes mellitus without complications’. Thus the clinical history of each patient can be uniquely represented by the concatenation of sequences of clinical descriptions $(\xi_{\theta_1}, \ldots ,\xi_{\theta_t})$, $\xi_{\theta_i} \in \Xi$, $i=1, \ldots, t$, ordered in time. Figure \ref{figure:hosp_gp} shows an example of fusing primary care and hospitalization records.

\footnotetext{Any patient data shown is simulated and does not represent data of real patients.}

\subsection{Label Generation: Oracle Feature Tagging for Disease Phenotyping}\label{labelling}
Given a set of clinical terms $\Theta$ and text descriptions $\Xi$, we rely on external oracles to assign labels to a given set of target phenotypes $\Delta$.  We assume that for each phenotype $d\in\Delta$ there exists a mapping \\ $\mathbf{1}_{d}:\Theta \times \Xi \rightarrow \{0,1\}$,
$(\theta, \xi_{\theta}) \mapsto \delta_{d}$ indicating whether the presence of $d$ can be inferred from the clinical term and its description. An aggregated phenotype label of $1$ is assigned, if $\mathbf{1}_{d}(\theta, \xi_{\theta})=1$ for any of the term-description pairs $(\theta, \xi_{\theta})$ in the input sequence, and $0$ otherwise (Figure \ref{figure:labelling}). Here, $\Delta$ is a set of disease phenotypes, that can be taken from disease-specific phenotyping algorithms \citep{10.1093/gigascience/giab059}, such as CALIBER \citep{kuan2019chronological}, which consists of manually-curated sets of clinical terms across primary and secondary care ontologies for defining 308 chronic and acute disease phenotypes.


\subsection{Model Design}

\begin{figure}[h!]
    \centering
    \includegraphics[width=14cm]{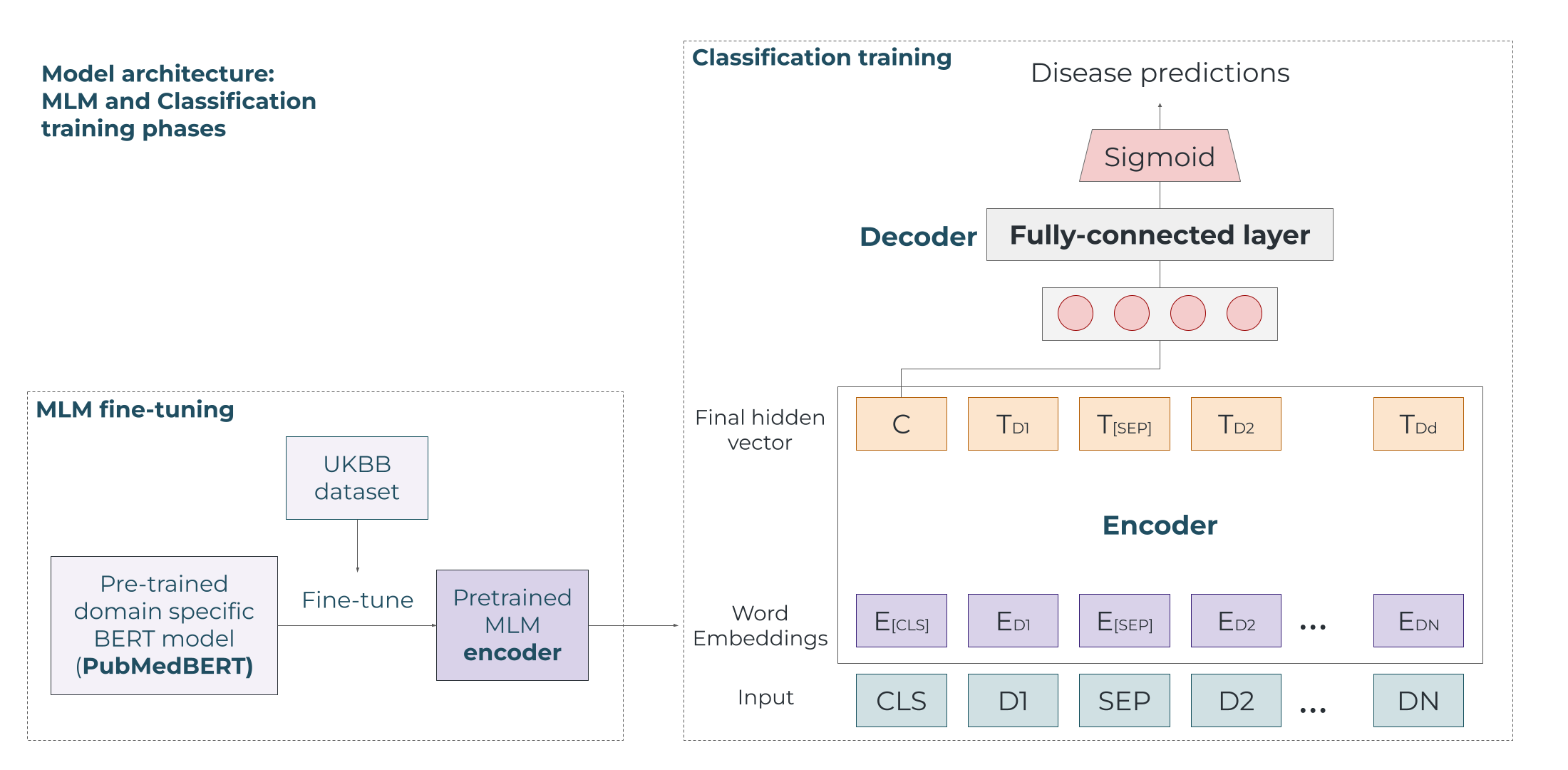}
    \caption{Diagram of the sEHR-CE model. The model is fine-tuned on the MLM task. We then use the pre-trained encoder and train  sEHR-CE. The input tokens are first encoded and the hidden vector of the [CLS] token is passed to the decoder, a fully connected linear layer. The output is passed through a sigmoid function to generate probabilities for each phenotype.}
\label{figure:model_diagram_small} 
\end{figure}

Figure \ref{figure:model_diagram_small} shows a diagram of sEHR-CE. 

\section{Experiment design and model predictions}\label{apd:model}




To assess the quality of our model predictions,  we chose four diseases that differ in terms of prevalence and clinical characteristics. Type 2 diabetes mellitus (T2DM) is one of the most prevalent chronic diseases worldwide \citep{t2d_who}, and patients are primarily diagnosed and managed in primary care (GP) \citep{diabetes_treatment}. Heart failure (HF) is one of the main causes of death in the older population \citep{heart_failure_roadmap}, and its acute manifestations are treated in hospital care. Malignant neoplasms of the breast and of the prostate are both less prevalent diseases almost exclusively present in only biologically females or males, respectively \citep{breast_cancer_incidence,prostate_epid}. We test the performance of our model on its ability to diagnose known cases, compare it to other methods, and evaluate associations of clinical features with the predictions on T2DM with available orthogonal data.

\subsection{Data Processing}\label{apd:data_proc}

The UK Biobank (UKBB) \citep{10.1371/journal.pmed.1001779} is a large-scale research study of around $500$k individuals between the ages of $40$ and $54$ at the time of recruitment. It includes rich genotyping and phenotyping data, both taken at recruitment and during primary and secondary care encounters (GP and hospital). We use patient records from GP and hospital visits in the form of code ontologies Read version2/ Clinical Terms Version 3, and ICD-9/10 together with their textual descriptions. We restrict the data set to individuals that have entries in both hospital and GP records, which reduces our cohort to $154,668$ individuals.  Requiring individuals to have entries in their GP records reduces bias towards acute events that usually present in hospitals, but we note that removing individuals without any hospital records may still bias the data towards more severe cases.  
We use CALIBER, previously validated phenotype definitions from \citet{kuan2019chronological} to label patients with T2DM, HF, malignant neoplasm of the breast, and malignant neoplasm of the prostate.

A patient can be admitted to the hospital for multiple days. We treat an entire hospital admission as one point in time  using the admission date, and only keep unique {ICD-10/ICD-9} codes for each visit. We aggregate visits that are less than a week apart into one visit keeping only unique codes. This approach removes repeated codes, thus avoiding redundancy and reducing sequence length. 

Only patients with at least $5$ clinical terms present in their clinical history are included to allow for sufficient information for any predictions, reducing the cohort to a final $129,932$ individuals. 
We use phenotype definitions from  CALIBER \citep{kuan2019chronological} to label patients with T2DM, HF, malignant neoplasm of the breast, and malignant neoplasm of the prostate. Each phenotype definition consists of a list of ICD10 and Read2, Read3 codes and their children. 

The usage of PubMedBERT restricts the length of input sequences we can use. To avoid excluding relevant clinical information by truncating the input sequences, we break up patient histories longer than the limit into multiple input sequences of smaller length with the same target vector.

\subsection{Model Training}
\label{apd:training}
\begin{figure*}[h]
\centering
\setlength{\belowcaptionskip}{-10pt}
\includegraphics[width=12cm]{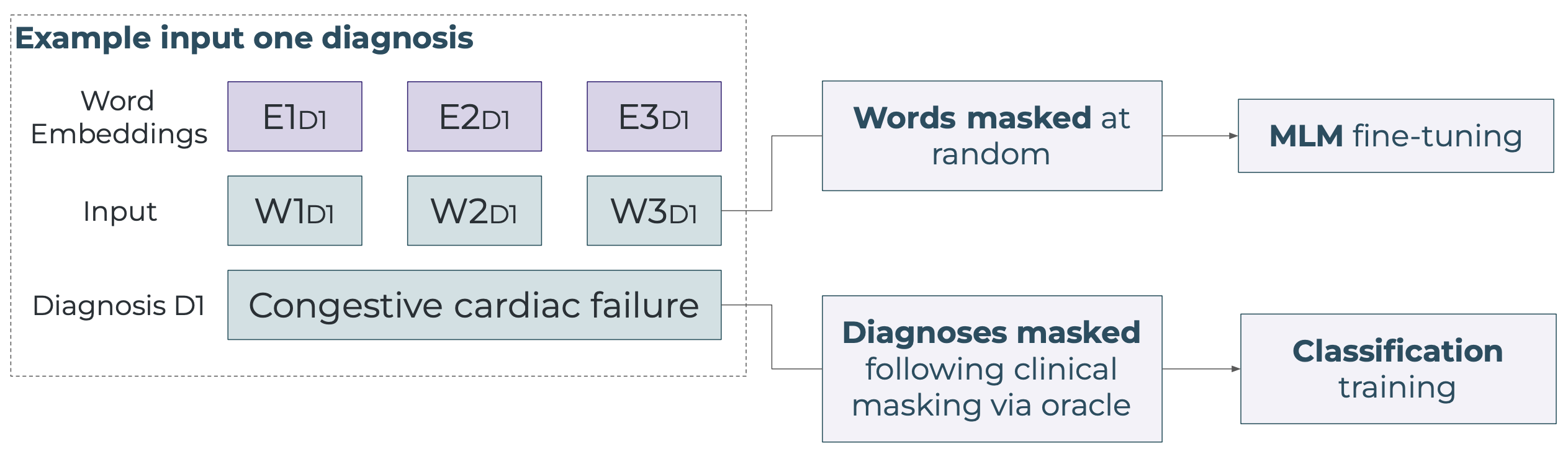}
\caption{Example of input diagnosis and different masking strategies for MLM fine-tuning and classification training. A description is encoded with multiple word embeddings. For MLM fine-tuning, words are masked at random; for classification training, whole descriptions are masked using clinical masking strategy described in Section \ref{data_augmentation}.}
\label{figure:input_diagram} 
\end{figure*}

We use the pre-trained language model PubMedBERT \citep{pubmedbert} as the encoder of the tokenised input sequences of clinical term descriptions. Since our input systematically differs from the general scientific text on which PubMedBERT was trained, we fine-tuned on the masked-language modeling (MLM) task (Figure \ref{figure:model_diagram_small}, Figure \ref{figure:input_diagram}), by masking words (e.g. descriptions) at random following the original BERT paper \citep{DBLP:journals/corr/abs-1810-04805}. The model, fine-tuned using the full UKBB cohort of $138,079$ patients, was trained with early stopping for $5$ epochs with a batch size of $32$ and a learning rate of $4\times10^{-5}$ using gradient descent with an AdamW optimizer, and weight decay of $0.01$. The output dimension of the encoder was $768$. 


The proposed model sEHR-CE is using the fine-tuned encoder and a fully connected linear layer as the decoder. To train on the multi-label classification task of outcome prediction, we split our data set into five equally sampled folds ${f_0,...f_4}$ containing unique patients, using a stratified sampling method to maintain the same phenotype proportion in every split \citep{Sechidis2011OnTS}, and mask the data according to our strategy (Section \ref{data_augmentation}, Figure \ref{figure:input_diagram}).

\begin{figure}[ht]
\centering
\includegraphics[width=8cm]{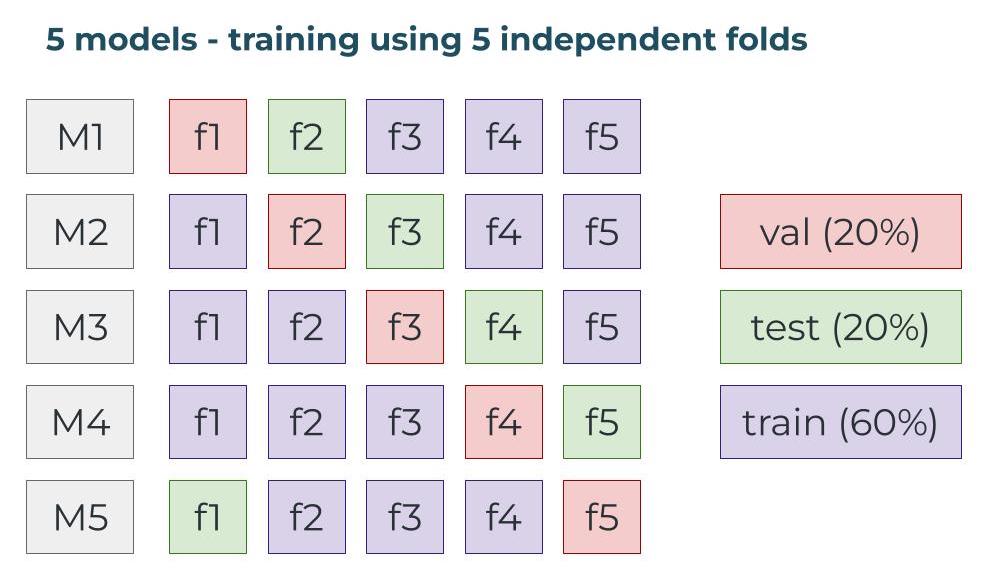}
\caption{Five models trained using five independent folds from the total data set using stratified sampling.}
\label{figure:folds_diagram} 
\end{figure}

We train a total of five models for $3$ epochs on three folds, holding back folds $f_i$ for validation and $f_{(i+1)\bmod{5}}$ for testing for model $i$, $i=1,\ldots,5$ (Figure \ref{figure:folds_diagram}). This maintains a $60/20/20$ training, validation and testing split overall while providing us with enough training and testing examples. All results presented are predictions of each model on its respective independent test set. 
We used a learning rate of $10^{-5}$, and a warm-up proportion of $0.25$. Performance was monitored every $0.25$ epochs on the validation fold. 

\subsection{Model evaluation}\label{apd:eval} 
As explained in \ref{sec:results}, we compare the performance of our model sEHR-CE to BEHRT \citep{https://doi.org/10.48550/arxiv.1907.09538}. BEHRT takes a tokenised sequence of clinical terms, age and position embeddings as input. Ontologies from hospital and GP records are mapped to CALIBER definitions \citep{kuan2019chronological}, removing unmapped terms. Phenotype definitions in CALIBER include different categories (for example, `diabetes' contains categories `type 1' and `type 2'), that were ignored by the original BEHRT publication, so we expanded the token set to define a token per CALIBER phenotype and category. A transformer model is pre-trained to predict masked tokens before it is trained to predict a set of possible diagnoses an individual may develop given the input sequence. Similarly, we trained five sEHR-CE models restricted to CALIBER tokens (denoted sEHR-CE-codes) for comparison. All results presented are predictions of each model on its respective independent test set. 

\begin{figure*}
\includegraphics[width=\textwidth]{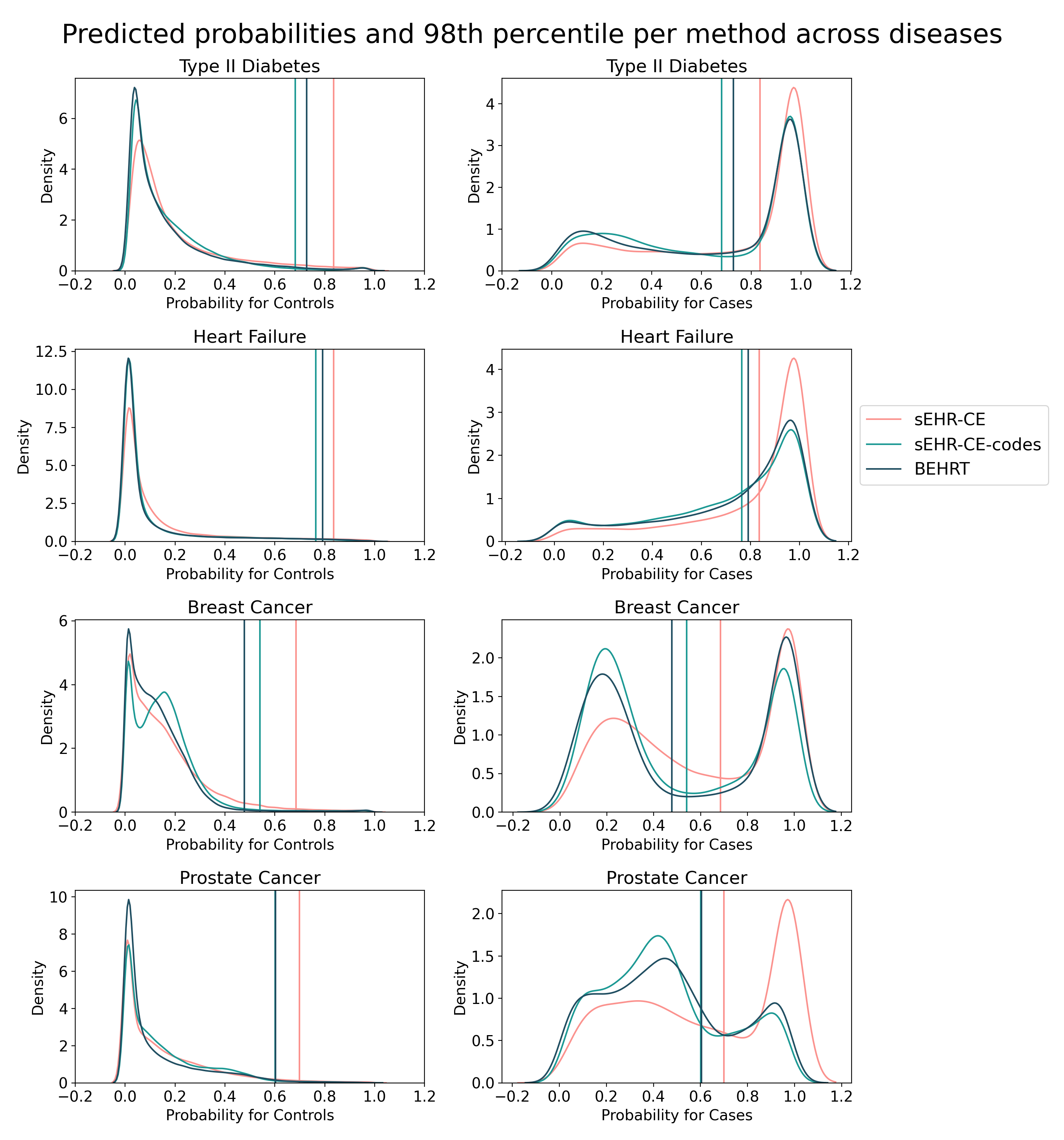}
\caption{Predicted probabilities for cases and controls in test sets across all phenotypes. Vertical lines indicate the $98$th percentile threshold.}
\label{figure:disease_dist_all} 
\end{figure*}

\begin{figure}[h!]
\centering
\includegraphics[width=\textwidth]{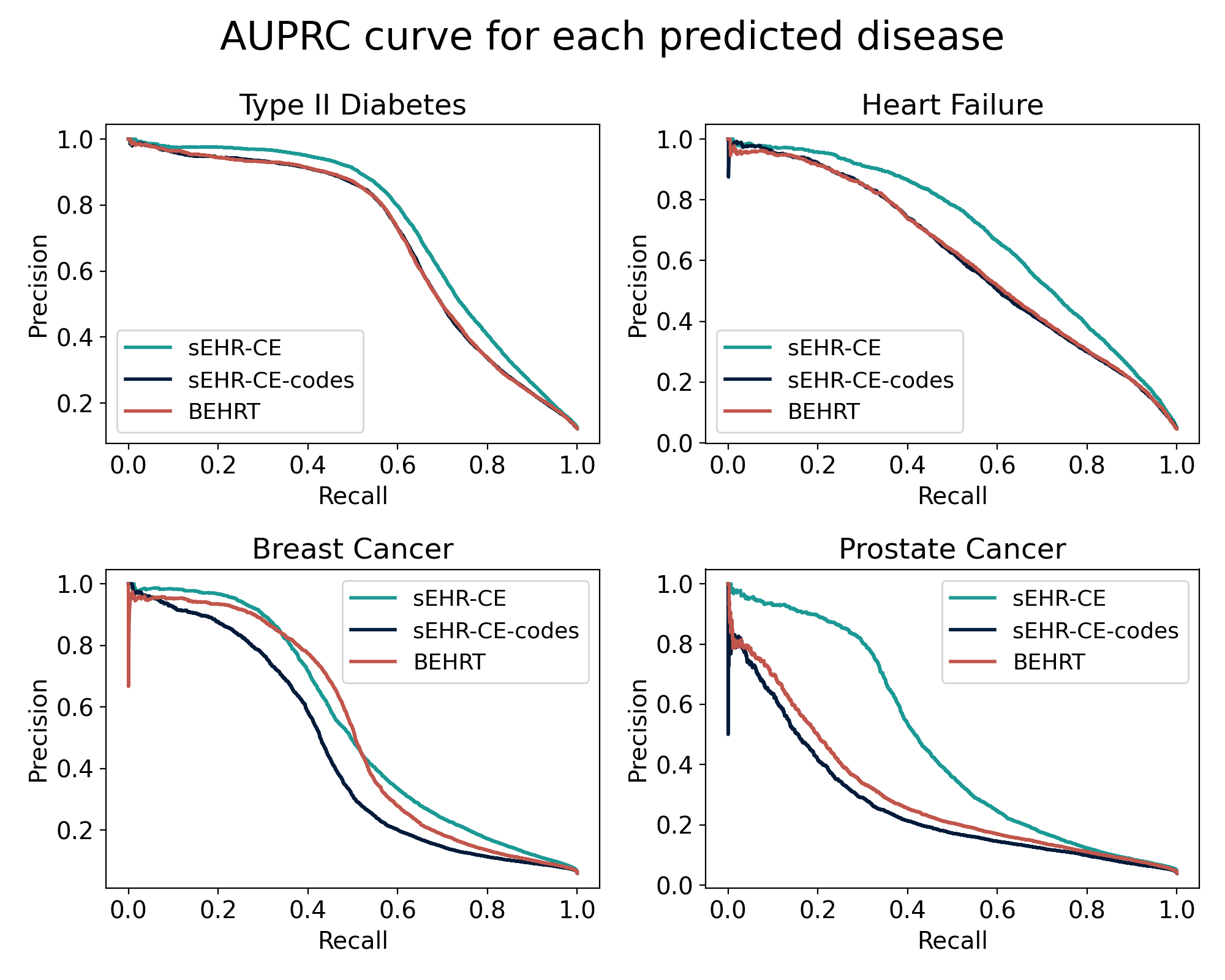}
\caption{AUPRC curves for each phenotype in the test sets.}
\label{figure:auprc} 
\end{figure}

Figure \ref{figure:disease_dist_all} shows the predicted probabilities for cases and controls across all phenotypes and models, and figure \ref{figure:auprc} shows the AUPRC curves for each phenotype and method in the test sets. To avoid inclusion of too many false positives, we defined missing cases as those controls with predicted probability in the $98$th percentile. 

\subsection{Expanded Qualitative Evaluation of T2DM Diagnosis Prediction}\label{apd:t2dm_eval}
Five groups were used to evaluate the model's T2DM diagnosis predictions. Table \ref{table:patient_groups} shows the percentiles of sEHR-CE's predicted probabilities to define each group, along with each size. 

\begin{table}[ht]
\centering
\small
\begin{tabular}{l|c}
\hline
Patient group                                                                                                  & Size   \\ \hline
Cases                                                                                                          & 16431  \\ \hline
Controls                                                                                                       & 113501 \\ \hline
\begin{tabular}[c]{@{}l@{}}Controls with high probability  \\  (p\textgreater{}=0.85, 98th percentile)\end{tabular} & 2020   \\ \hline
\begin{tabular}[c]{@{}l@{}}Cases with high  probability \\  (p\textgreater{}=0.985, 90th percentile)\end{tabular}    & 2072   \\ \hline
\begin{tabular}[c]{@{}l@{}}Cases with low  probability \\ (p\textless{}=0.25, 12th percentile)\end{tabular}        & 2343  \\\hline
\end{tabular}
\caption{Case and control cohorts and groups of interest based on sEHR-CE's predicted probability for T2DM in the test sets.}
\label{table:patient_groups} 
\end{table}

We then investigated the association of predicted probability and several proxies of disease severity: number of GP and hospital admissions, survival and risk of cardiovascular disease. 

\newpage
\newpage
\subsubsection{Number of GP visits and hospital admissions}\label{number_admissions}
\begin{figure}[!h]
\centering
\includegraphics[width=\textwidth]{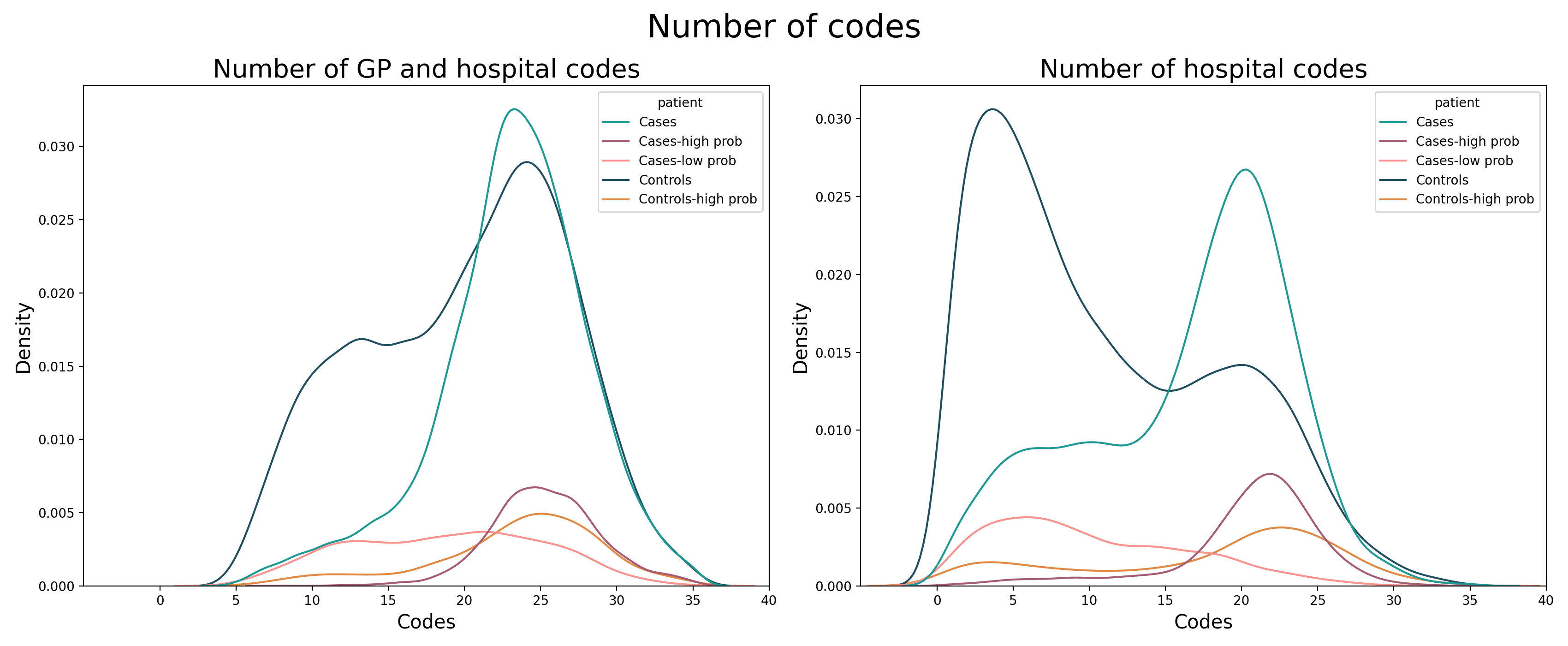}
\caption{Distribution of number of hospital and GP codes per patient group. Matched controls on sex.}
\label{figure:number_codes} 
\end{figure}

As expected, both cases and controls with a high predicted probability of a T2DM diagnosis, exhibit a slightly higher number of GP and hospital visits than the other groups (Figure  \ref{figure:number_codes}), indicating that they are experiencing a more severe form of T2DM requiring care. This is particularly higher in the case of hospital visits, indicating patients experiencing acute events: both cases and controls with a high predicted probability visit a hospital approximately $10$ times more often than their low probability counterparts. 

Although the model was not given information from which data source the input data was coming from, this analysis indicates that it has learned to associate acute events with disease severity. 

\subsubsection{Survival analysis}\label{surivial_analysis}
To compare survival across different groups of individuals, we use the Kaplan-Meier estimator with all-cause mortality as the endpoint with right-censored data accounting for individuals without any event occurrence since the last follow-up. Both cases and controls with high predicted probabilities had the lowest survival, followed by general cases, controls and finally cases with low predicted probability (Figure \ref{figure:survival}), indicating that the model's predicted probability is associated with survival. 


\subsubsection{Cardiovascular Risk}\label{cariovascular_risk}
\begin{figure}
    \centering
    \begin{subfigure}[t]{0.49\textwidth}
        \includegraphics[width=\textwidth]{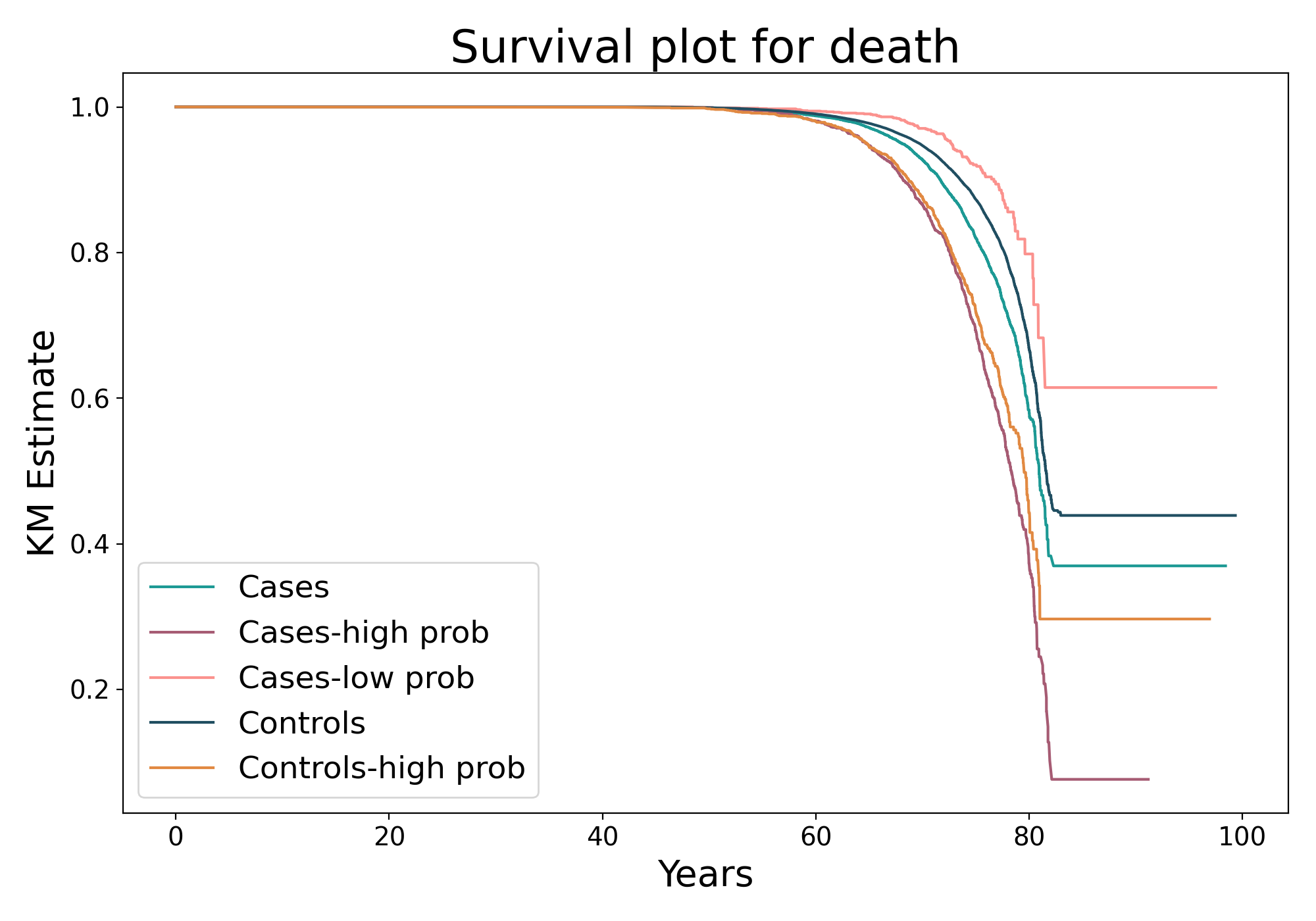}
    \caption{}
    \label{figure:survival}
    \end{subfigure}
    \begin{subfigure}[t]{0.49\textwidth}
        \includegraphics[width=\textwidth]{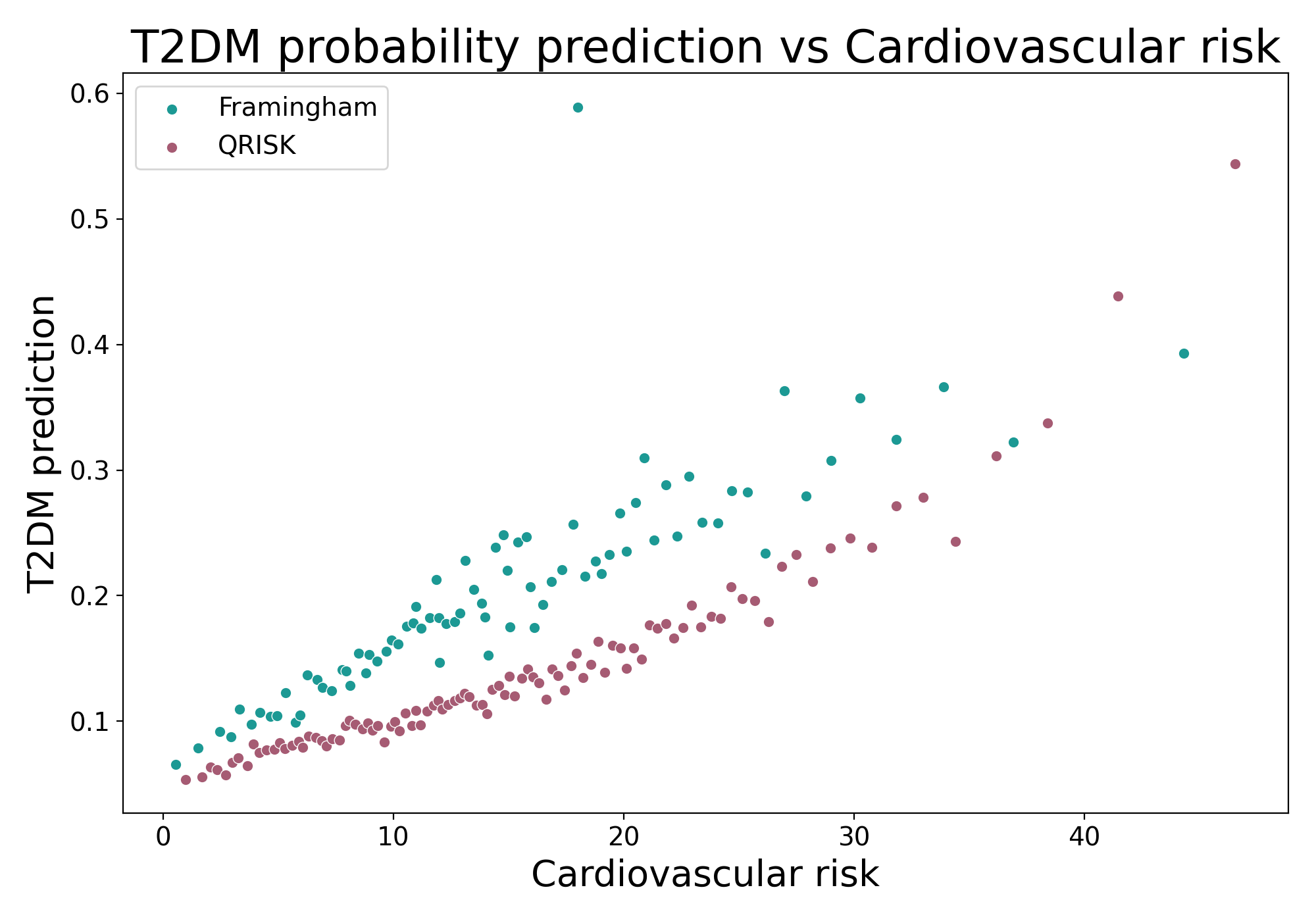}
        \caption{}
        \label{figure:cardio_risk_scatter}
    \end{subfigure}
    \caption{\ref{figure:survival} Death survival plots for different patient groups. \ref{figure:cardio_risk_scatter} Framingham and QRISK cardiovascular scores vs T2DM probability prediction. Each point represents the median of each percentile of cardiovascular risk.}
\end{figure}

T2DM is a known risk factor and comorbidity of cardiovascular disease, which, in turn, is the most prevalent cause of death in T2DM patients. The GP records contain Framingham and QRISK3 scores; these are two scores that assess an individual's risk of developing cardiovascular disease  within the next 10 years, based on several coronary risk factors. The Framingham score is derived from an individual's age, gender, total cholesterol, high-density lipoprotein cholesterol, smoking habits, and systolic blood pressure, whereas the QRISK score, which is almost exclusively used now, extends this score with additional factors such as body mass index, ethnicity, measures of deprivation, chronic kidney disease, rheumatoid arthritis, atrial fibrillation, diabetes mellitus, and anti-hypertensive treatment. Both cases and controls with high predicted probability of having T2DM had a higher risk of developing cardiovascular disease compared to their low predicted probability counterparts (Figure \ref{figure:cardio_risk_scatter}) indicating that the model has learned to associate the risk of developing both diseases at the same time. 




\end{document}